\documentclass[conference,letterpaper]{IEEEtran}

\usepackage[utf8]{inputenc} 
\usepackage[T1]{fontenc}
\usepackage{url}
\usepackage{ifthen}
\usepackage{cite}
\usepackage[cmex10]{amsmath} 
\usepackage{amssymb,amsfonts}
\usepackage{mathtools}

\newtheorem{lemma}{Lemma}

\newtheorem{theorem}{Theorem}
\usepackage{algorithmic}
\usepackage{graphicx}
\usepackage[caption=false,font=footnotesize]{subfig}
\usepackage{textcomp}
\usepackage{xcolor}

\begin{document}
\title{An Efficient Difference-of-Convex Solver for Privacy Funnel} 


\author{%
   \IEEEauthorblockN{Teng-Hui~Huang and Hesham~El Gamal}
   \IEEEauthorblockA{School of Electrical and Computer Engineering, University of Sydney\\
                     NSW, Australia\\
                    \{tenghui.huang, hesham.elgamal\}@sydney.edu.au
                   }
}


\maketitle


\begin{abstract}
We propose an efficient solver for the privacy funnel (PF) method, leveraging its difference-of-convex (DC) structure. The proposed DC separation results in a closed-form update equation, which allows straightforward application to both known and unknown distribution settings. For known distribution case, we prove the convergence (local stationary points) of the proposed non-greedy solver, and empirically show that it outperforms the state-of-the-art approaches in characterizing the privacy-utility trade-off. The insights of our DC approach apply to unknown distribution settings where labeled empirical samples are available instead. Leveraging the insights, our alternating minimization solver satisfies the fundamental Markov relation of PF in contrast to previous variational inference-based solvers. Empirically, we evaluate the proposed solver with MNIST and Fashion-MNIST datasets. Our results show that under a comparable reconstruction quality, an adversary suffers from higher prediction error from clustering our compressed codes than that with the compared methods. Most importantly, our solver is independent to private information in inference phase contrary to the baselines.
\end{abstract}

\section{Introduction}
The privacy funnel (PF) method attracts increasing attention recently, driven by the need for enhanced data security as modern machine learning advances. However, due to the non-convexity of the involved optimization problem, the strict Markov constraint, and the requirement of full knowledge of a probability density, the PF problem is well-known to be difficult to solve. In the PF method, the goal is to solve the following optimization problem~\cite{privacyF2014}:
\begin{IEEEeqnarray}{rCl}
    \underset{P(Z|X)\in\Omega_{Z|X}}{\min}\,&& I(Z;Y),\nonumber\\
    \text{subject to}\,&& I(Z;X)\geq \eta_X,\IEEEeqnarraynumspace\IEEEyesnumber
    \label{eq:privacy_funnel_problem}
\end{IEEEeqnarray}
where $\eta_X$ denotes a control threshold. In the above problem, the joint distribution $P(X,Y)$ is assumed to be known and the random variables satisfy the Markov chain $Y\rightarrow X\rightarrow Z$. Here, $Y$ is the private information; $X$ the public information and $Z$ is the released (compressed) codes; $I(Z;X)$ denotes the mutual information between two random variables $Z$ and $X$. In~\eqref{eq:privacy_funnel_problem}, the variables to optimize with are the stochastic mappings $P(Z|X)$ with $\Omega_{Z|X}$ denotes the associated feasible solution set. The aim of the problem is to characterize the fundamental trade-off between the privacy leakage and the utility of the released information. One of the challenges in solving~\eqref{eq:privacy_funnel_problem} is the non-convexity of the problem. It is well-known that the mutual information $I(Z;X)$ with known marginal $P(X)$ is a convex function with respect to $P(Z|X)$~\cite{cover1999elements}; Additionally, due to the Markov relation, $I(Z;Y)$ is a convex function of $P(Z|X)$ as well. Together, \eqref{eq:privacy_funnel_problem} minimizes a convex objective under a non-convex set. Previous solvers for PF address the non-convexity by restricting the feasible solution set further for efficient computation. Both~\cite{privacyF2014} and~\cite{submodpf2019} consider discrete settings and focus on deterministic $P(z|x)$, i.e., a realization of $x\in\mathcal{X}$ belongs to a single cluster $z\in\mathcal{Z}$ only. They propose greedy algorithms to solve the reduced problem and differ in choosing the subset of $\mathcal{Z}$ to combine with. However, the imposed restriction limits the characterization of the privacy-utility trade-off. In our earlier work~\cite{huang2022linear}, we propose a non-convex alternating direction methods of multipliers (ADMM) solver for PF which is non-greedy, but it suffers from slow convergence~\cite{attouch2010proximal}.

Another challenge of PF is knowing the joint distribution $P(X,Y)$. Overcoming this is important for practical application of the method as $P(X,Y)$ is prohibitively difficult to obtain in general. Nonetheless, it is relatively easy to obtain samples of the joint distribution, which become a training dataset as an indirect access to $P(X,Y)$. Recently, inspired by the empirical success of information theoretic-based machine learning, several works have attempted to extend the PF method through the theory of variational inference~\cite{alemi2016deep}. In~\cite{rodriguez2021variational}, the conditional mutual information form of PF due to $I(Y;Z)=I(X;Z)-I(X;Z|Y)$ is adopted, whose variational surrogate bound is estimated with empirical samples. In \cite{razeghi2024deep} a similar formulation is adopted, but they restrict the parameter space and implement a generative deep architecture for interchangeable use of the empirical and the generated $Y$ samples for estimation. However, owing to the dependency of the private information $Y$ in estimating $I(X;Z|Y)$, these previous solvers violate the fundamental Markov relation in PF method. Consequently, samples of $X,Y$ are required for their solvers which means supervision is required in both model training and inference phases.

Different from the previous works, we focus on more computationally efficient PF solvers for both known and unknown (empirical samples available) joint distribution settings based on the difference-of-convex (DC) structure of~\eqref{eq:privacy_funnel_problem}. In known joint distribution settings, inspired by the difference-of-convex algorithm (DCA), we propose a novel PF solver with closed-form update equation. In our evaluation, the proposed solver not only covers solutions obtained from the state-of-the-art solvers but also attains non-trivial solutions that the compared greedy solvers are infeasible to reach. In unknown $P(X,Y)$ settings, our DC method complies with the fundamental Markov relation, therefore requires public information $X$ only in inference phase in contrast to previous approaches. We implement our solver with deep neural networks and evaluate the model on MNIST and Fashion-MNIST datasets. In all evaluated datasets, our approach outperforms the state-of-the-art. Specifically, under a comparable reconstruction quality of the public information $X$, an adversary's clustering accuracy (leakage of the private information $Y$) with the released codes $Z$ of ours is significantly lower than the compared approaches.

\paragraph*{Notations}
Upper-case letters denote random variables (RVs) and lower-case letters are their realizations. The calligraphic letter denotes the sample space, e.g., $x\in\mathcal{X}$ with cardinality $|\mathcal{X}|$. The vectors and matrices are denoted in bold-face symbols, e.g.,  $\boldsymbol{x},\boldsymbol{X}$ respectively. We denote the conditional probability mass function as a long vector:
\begin{IEEEeqnarray}{rCl}
    \boldsymbol{p}_{z|x}:=\begin{bmatrix}P(x_1|z_1)&P(x_1|z_2)&\cdots & P(x_{|\mathcal{X}|}|z_{|\mathcal{Z}|})\end{bmatrix}^T.
\end{IEEEeqnarray}
The Kullback-Leibler divergence of two measurable proper densities $\mu,\nu$ is denoted as $D_{KL}[\mu\parallel \nu]$~\cite{cover1999elements}. 

\section{A Difference-of-Convex Algorithm for Privacy Funnel}
For simplicity, we focus on discrete random variables settings in this section. We start with a summary of DC programming and refer to the excellent review~\cite{le2018dc} for the development and recent advances. For a DC program, the canonical form is given by:
\begin{IEEEeqnarray}{rCl}
    w^*=\underset{w\in\mathcal{W}}{\arg\min}\,f(w)-g(w),\IEEEeqnarraynumspace\IEEEyesnumber\label{eq:dc_program}
\end{IEEEeqnarray}
where $f,g$ are functions of $w$, and $\mathcal{W}$ denotes the feasible solution set of $w$. \eqref{eq:dc_program} relates to the PF method through rewriting~\eqref{eq:privacy_funnel_problem} as an unconstrained PF Lagrangian:
\begin{IEEEeqnarray}{rCl}
    \mathcal{L}_{PF}:=I(Z;Y)-\beta I(Z;X),\IEEEeqnarraynumspace\IEEEyesnumber\label{eq:pf_unconstrained}
\end{IEEEeqnarray}
where $\beta>0$ is a multiplier. The above is a DC program when $P(X,Y)$ is known for a fixed $\beta$. This is because $I(Z;X)$ is convex w.r.t. $P(Z|X)$ for known $P(X)$, and $-H(Z|Y)$ is convex w.r.t. $P(Z|X)$ as the Markov relation $P(Z|Y)=\sum_{x\in\mathcal{X}}P(Z|x)P(x|Y)$ for known $P(Y)$~\cite{cover1999elements}. The DC structure of~\eqref{eq:pf_unconstrained} can be leveraged for efficient optimization. By substituting the concave part with its first order approximation, we obtain a convex sub-objective. Then we iteratively minimize the convex surrogate sub-objective until the loss saturates. The monotonically decreasing loss values is assured due to the convexity of the sub-objective functions. However, the optimality of the converged solution is lost due to the non-convexity of the overall problem. There are well-developed efficient solvers for DC problems in literature, known as the Difference-of-Convex Algorithm (DCA). However, one of the key challenges in solving DC problems with DCA is that there exist infinitely many combinations of DC pairs~\cite{le2018dc}. Furthermore, a poorly chosen pair may result in slow convergence and numerical instability. We address the issues by proposing the following DC separation:
\begin{IEEEeqnarray}{rCl}
    f(\boldsymbol{p}_{z|x}):&=&-H(Z|Y),\IEEEeqnarraynumspace\IEEEyesnumber\label{eq:pf_dca_fg}\IEEEyessubnumber\label{eq:pf_dca_f}\\
    g(\boldsymbol{p}_{z|x}):&=&-H(Z)+\beta I(Z|X),\IEEEeqnarraynumspace\IEEEyessubnumber\label{eq:pf_dca_g}
\end{IEEEeqnarray}
Then we apply \eqref{eq:pf_dca_fg} to DCA:
\begin{IEEEeqnarray}{rCl}
    \boldsymbol{p}^{k+1}_{z|x}:=\underset{\boldsymbol{p}_{z|x}\in\Omega_{z|x}}{\arg\min}\, f(\boldsymbol{p}_{z|x})-\langle\nabla g(\boldsymbol{p}^{k}_{z|x}),\boldsymbol{p}_{z|x}-\boldsymbol{p}^{k}_{z|x}\rangle,\IEEEeqnarraynumspace\IEEEyesnumber\label{eq:DCA_for_pf}
\end{IEEEeqnarray}
where the superscript $k$ denotes the iteration counter. For known $P(X,Y)$ in discrete settings,~\eqref{eq:DCA_for_pf} can be expressed as a linear equation (see Appendix~\ref{appendix:derive_known_jt}):
\begin{IEEEeqnarray}{rCl}
    \boldsymbol{B}_{y|x} \log{\boldsymbol{p}_{z|y}} = \log{\boldsymbol{p}^k_z}+\beta\log{\frac{\boldsymbol{p}^k_{z|x}}{\boldsymbol{p}^k_z}},\IEEEeqnarraynumspace\IEEEyesnumber\label{eq:dca_update_lineareq}
\end{IEEEeqnarray}
where $\boldsymbol{B}_{y|x}:=\boldsymbol{I}_{|\mathcal{Z}|}\otimes \boldsymbol{Q}_{y|x}$, with $\boldsymbol{I}_{d}$ denotes the $d$-dimensional identity matrix, $\otimes$ the Kronecker product, and $\boldsymbol{Q}_{y|x}$ the matrix-form of $P(Y|X)$ whose $(i,j)$-entry is $P(y_i|x_j)$. 
We follow the assumption in private source coding literature that $\boldsymbol{B}_{y|x}$ is full-row rank~\cite{7282765,6942222,7541465,9678374,9272984}. Then we denote the Moore-Penrose pseudo-inverse of $\boldsymbol{B}_{y|x}$ as $\boldsymbol{B}^\dagger_{y|x}:=(\boldsymbol{B}_{y|x}^T\boldsymbol{B}_{y|x})^{-1}\boldsymbol{B}_{y|x}^T$. Note that due to the Markov relation, we have $\boldsymbol{A}_{x|y}\boldsymbol{p}_{z|x}=\boldsymbol{p}_{z|y}$ with $\boldsymbol{A}_{x|y}:=\boldsymbol{I}_{\mathcal{Z}}\otimes \boldsymbol{Q}_{x|y}$ defined similarly as $\boldsymbol{B}_{y|x}$. Summarizing the above, \eqref{eq:dca_update_lineareq} reduces to:
\begin{IEEEeqnarray}{rCl}
    \tilde{\boldsymbol{p}}_{z|y}\propto\exp\{\boldsymbol{B}^\dagger_{y|x}\boldsymbol{c}^k_{z|x}\},\IEEEeqnarraynumspace\IEEEyesnumber\label{eq:dca_raw_pzcy_dagger}
\end{IEEEeqnarray}
where $\boldsymbol{c}^k_{z|x}:=\log{\boldsymbol{p}^k_z}+\beta \log{\frac{\boldsymbol{p}^k_{z|x}}{\boldsymbol{p}^k_z}}$, $\tilde{\boldsymbol{p}}_{z|y}$ denotes the un-normalized probability vector and $\propto$ represents proportional to. After normalization, followed by the Markov relation, the insight from applying DCA to PF gives an update equation:
\begin{IEEEeqnarray}{rCl}
    \boldsymbol{A}_{x|y}\boldsymbol{p}_{z|x}=\boldsymbol{\phi}(\boldsymbol{B}_{y|x}^\dagger \boldsymbol{c}^k_{z|x}):=\boldsymbol{q}^k_{z|y},\IEEEeqnarraynumspace\IEEEyesnumber\label{eq:dca_inner_update}
\end{IEEEeqnarray}
where the $z\in\mathcal{Z}$ element is given by $\phi_z=e^z/\sum_{z'\in\mathcal{Z}}e^{z'}$. In practice, $\boldsymbol{\phi}$ is implemented as a softmax function.
The above implies that the update of $\boldsymbol{p}_{z|x}$ from step $k$ to step $k+1$ requires solving the linear equation \eqref{eq:dca_inner_update}. However, since $\boldsymbol{A}_{x|y}^\dagger$ might not exist ($\boldsymbol{B}_{y|x}$ is full-row rank), \eqref{eq:dca_inner_update} is an under-determined linear program over non-negative simplice. Heuristically, one would prefer a solution with the lowest complexity, which is often achieved by imposing $q$-norm (e.g., $q\in\{0,1,2\}$) constraints as regularization~\cite{donoho2006compressed}. Based on this intuition, we therefore solve the following sub-problem:
\begin{IEEEeqnarray}{rCl}
    \underset{\boldsymbol{p}_{z|x}\in\Omega_{z|x}}{\min}\,&& \lVert \boldsymbol{p}_{z|x}\rVert_q,\nonumber\\
    \text{subject to}\,&& \boldsymbol{A}_{x|y}\boldsymbol{p}_{z|x}=\boldsymbol{q}^k_{z|y},\IEEEeqnarraynumspace\IEEEyesnumber\label{eq:inner_update_opt_problem}
\end{IEEEeqnarray}
where $\lVert \boldsymbol{v}\rVert_q:=(\sum_{v\in\mathcal{V}}|v|^q)^{1/q}$. Similar to relaxing the ``hard'' constraints in classical regression problems for applying more efficient algorithms, we solve the following relaxed problem of \eqref{eq:inner_update_opt_problem} where a soft constraint is considered instead:
\begin{IEEEeqnarray}{rCl}
    \boldsymbol{p}^{k+1}_{z|x}:=\underset{\boldsymbol{p}_{z|x}\in\Omega_{z|x}}{\arg\min}\,&& \frac{1}{2}\lVert \boldsymbol{A}_{x|y}\boldsymbol{p}_{z|x}-\boldsymbol{q}_{z|y}^k\rVert^2_{2}+\alpha \lVert \boldsymbol{p}_{z|x}\rVert_q,\IEEEeqnarraynumspace\IEEEyesnumber\label{eq:lasso_pf}
\end{IEEEeqnarray}
where $\alpha>0$ is a relaxation coefficient. The relaxed problem tolerates an approximation error of the linear equation instead of a strict equality constraint. Remarkably, \eqref{eq:inner_update_opt_problem} relates to the classical regression problems such as Ridge regression~\cite{draper2002generalized}, LASSO~\cite{gaines2018algorithms}, and sparse recovery~\cite{pilanci2012recovery}. But in \eqref{eq:inner_update_opt_problem} and \eqref{eq:lasso_pf}, the feasible solution set $\Omega_{z|x}$ is the probability simplice. For $q=2$, since~\eqref{eq:lasso_pf} corresponds to a convex optimization problem, each step can be solved efficiently with off-the-shelf ridge regression solvers~\cite{scikit-learn}. As for $q=1$ (convex surrogate for $q=0$), however, the LASSO solver~\cite{gaines2018algorithms} does not work well. This is because $
    \lVert \boldsymbol{p}_{z|x}\rVert_1=\sum_{x\in\mathcal{X}}|\boldsymbol{p}_{Z|x}|=|\mathcal{X}|$, since $\sum_{z\in\mathcal{Z}}P(z|x)=1,\forall x\in\mathcal{X}$ always. To address this, we provide the following alternative for sparse recovery ($q=0$) with details referred to Appendix \ref{appendix:kn_sparse_problem}:
\begin{IEEEeqnarray}{rCl}
    l^{k+1}_{z|x}:=\underset{l_{z|x}\in \mathcal{H}_{z|x}}{\arg\min}\, \frac{1}{2}\lVert {lse}_{x}(l_{x|y}+l_{z|x})-l^k_{\phi}\rVert^2+\alpha\lVert l_{z|x}\rVert_1,\IEEEeqnarraynumspace\IEEEyesnumber\label{eq:lse_q1}
\end{IEEEeqnarray}
where $l_{x|y}:=\log{\boldsymbol{p}_{x|y}}$, $lse(x):=\log\sum_{x\in\mathcal{X}}\exp\{x\}$ is the log-sum-exponential function; $\mathcal{H}_{z|x}$ is the feasible set for $l_{z|x}\in[-M,-m]$ for some $M>m>0$ such that $\sum_{z\in\mathcal{Z}}\exp\{l_{z|x}\}=1,\forall x\in\mathcal{X}$; the subscript of $\text{lse}$ denotes the variable for summation, and $l_\phi^k:=\log{\boldsymbol{\phi}(\boldsymbol{B}_{y|x}^\dagger \boldsymbol{c}^k_{z|x})}$. Then after solving~\eqref{eq:lse_q1}, we can project the obtained $l_{z|x}$ to a feasible solution $\boldsymbol{p}_{z|x}$ through a softmax function: $\boldsymbol{p}_{z|x}=\text{Softmax}(l_{z|x})$.

For both implementations, the proposed solver guarantees convergence to a local stationary point.
\begin{theorem}\label{thm:conv_discrete_kn}
    For both implementations~\eqref{eq:lasso_pf} ($q=2$) and \eqref{eq:lse_q1}, the sequence $\{\boldsymbol{A}_x\boldsymbol{p}_{z|x}^k\}_{k\in\mathbb{N}}$, obtained from~\eqref{eq:DCA_for_pf}, converges to a stationary point $\boldsymbol{A}_x\boldsymbol{p}_{z|x}^*$ such that $\nabla f(\boldsymbol{p}^*_{z|x})=\nabla g(\boldsymbol{p}^*_{z|x})$ with $f,g$ defined in~\eqref{eq:pf_dca_fg}, and $\boldsymbol{A}_x:=\boldsymbol{I}_{\mathcal{Z}}\otimes \boldsymbol{Q}_x$.
\end{theorem}
\begin{IEEEproof}
    See Appendix \ref{appendix:proof_thm}.
\end{IEEEproof}

\section{Extension to Unknown Distributions}

The main difficulty in applying PF is the full knowledge of the joint distribution $P(X,Y)$~\cite{privacyF2014}. In \eqref{eq:lasso_pf}, this corresponds to defining the operators $\boldsymbol{A}_{x|y}$ and $\boldsymbol{B}_{y|x}^{\dagger}$. Without knowing $P(X,Y)$, these operators are intractible. To address this, we leverage the DC structure of the PF problem again, but now take expectation with respect to $P(Z,X)=P(Z|X)P(X)$ on the update equation (see Appendix~\ref{appendix:derive_unkn_jt}):
\begin{IEEEeqnarray}{rCl}
    I(Z;Y)+D_{KL}[P_z\parallel P_z^k]-\beta\mathbb{E}_{z,x}\left[\log{P^k(X|Z)}\right]=0,\IEEEeqnarraynumspace\IEEEyesnumber\label{eq:expect_dca_update}
\end{IEEEeqnarray}
where the superscript $k$ denotes the iteration counter. The above update equation implies that the step $k+1$ solution $P^{k+1}(Z|X)$ is obtained from solving \eqref{eq:expect_dca_update}, given the step $k$ solution $P^k(Z|X)$. 

The expectation form~\eqref{eq:expect_dca_update} is useful because the computationally prohibitive knowledge of $P(X,Y)$ can be approximated efficiently through the theory of the variational inference, followed by Monte-Carlo sampling~\cite{kingma2013auto,alemi2016deep}. The derived surrogate bound involves an auxiliary variable, which corresponds to the released information $Z$, and is associated with a variational distribution to be designed. For the mutual information $I(Z;Y)=H(Y)-H(Y|Z)$:
\begin{IEEEeqnarray}{rCl}
    H(Y|Z)
    \leq \mathbb{E}_{y,z;\phi}\left[\log{\frac{1}{Q_\phi(Y|Z)}}\right],\IEEEeqnarraynumspace\IEEEyesnumber\label{eq:vi_ent_yzc}
\end{IEEEeqnarray}
where the above surrogate bound is tight when $Q_\phi(Y|Z)=P(Y|Z)$. As for the other terms in~\eqref{eq:expect_dca_update}, we parameterize $P_\theta(Z)=\mathcal{N}(\boldsymbol{0},\text{diag}(\boldsymbol{\sigma}_\theta^2))$ as standard Gaussian and $P_\phi(X|Z)=\mathcal{N}(\boldsymbol{\mu}_{\phi}(\boldsymbol{z}),\boldsymbol{I})$ as conditional Gaussian distributions. This transforms the intractible problem into standard parameter estimation. Also, we parameterize the encoder $P_\theta(Z|X)$, corresponding to the expectation operator in~\eqref{eq:expect_dca_update}, through the evident lower bound technique as adopted in the variational autoencoders (VAE)~\cite{kingma2013auto}. Here, we have $P_\theta(Z|X)\sim\mathcal{N}(\boldsymbol{\mu}_\theta(\boldsymbol{x}),\boldsymbol{\Sigma}_\theta(\boldsymbol{x}))$. Combining the above, we propose the following alternating solver:
\begin{IEEEeqnarray}{rCl}
    \phi^{*}:=\underset{\phi\in\Phi}{\arg\min}\,&& -\beta \mathbb{E}_{x,z;\theta^k}\left[\log{P_\phi(X|Z)}\right]\nonumber\\&&-\mathbb{E}_{y,z;\theta^k}\left[\log{Q_\phi(Y|Z)}\right],\IEEEyesnumber\label{eq:unkn_alg}\IEEEeqnarraynumspace\IEEEyessubnumber\label{eq:unkn_alg_step1}\\
    \theta^{k+1}:=\underset{\theta\in\Theta}{\arg\min}\,&&\frac{1}{2}\lVert I_{\phi^*;\theta}(Z;Y)+D_{KL}[P_{z;\theta}\parallel P_{z;\theta^k}]\nonumber\\
    &&-\beta\mathbb{E}_{z,x;\theta}\left[\log{P_{\phi^*}}(X|Z)\right]\rVert_2^2\nonumber\\
    && +\alpha \mathbb{E}_x\left[D_{KL}[P_{z|x;\theta}\parallel r_Z]\right],\IEEEeqnarraynumspace\IEEEyessubnumber\label{eq:unkn_alg_step2}
\end{IEEEeqnarray}
where $r_Z=\mathcal{N}(\boldsymbol{0},\boldsymbol{I})$ is a reference probability density function, regularizing the encoder; and $\alpha>0$ is a control threshold. Essentially, we alternate between fitting the empirical samples of $X,Y$ \eqref{eq:unkn_alg_step1} and solving the DCA update sub-problem with an extra regularization term \eqref{eq:unkn_alg_step2}. The two steps are repeated until a pre-determined number of iterations is reached.

The optimization of \eqref{eq:unkn_alg} is efficient. Each of the two KL divergence terms are computed between two Gaussian densities which has closed-form expression~\cite{duchi2007derivations}; the parameterized $P_\phi(X|Z)$ and $Q_\phi(Y|Z)$ (hence $I(Z;Y)\approx H(Y)-H_\phi(Y|Z)$) can be estimated with mean-squared error (MSE) and a categorical cross entropy loss respectively. Finally, two remarks are in order:
\begin{itemize}
    \item While variational inference is adopted of the proposed solver, but we apply it on the expectation form of the proposed DC update equation~\eqref{eq:expect_dca_update}, not on the PF Lagrangian nor on $I(X;Z|Y)$~\cite{rodriguez2021variational,razeghi2024deep}.
    \item The parameterization of~\eqref{eq:unkn_alg} can be extended to other density functions such as Laplace, Bernoulli and exponential distributions~\cite{park2019variational,huang2023Wynerfinger}.
\end{itemize}

\section{Evaluation}
\begin{figure*}[!ht]
    \centering
    \subfloat[DCA \eqref{eq:lse_q1} with $q=1$ versus \textit{Submodular}~\cite{submodpf2019}]{
        \label{subfig:discrete_q1_vs_sub}
        \includegraphics[width=2.9in]{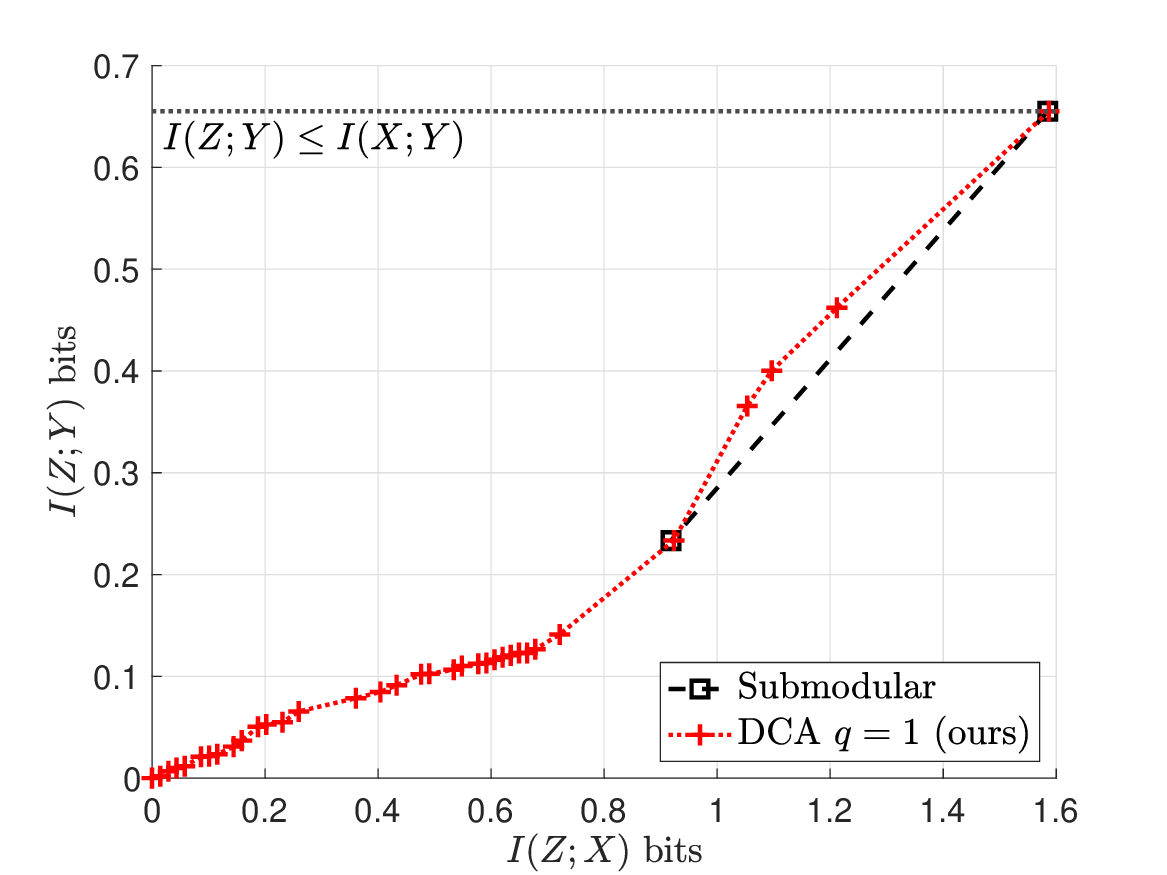}
    }
    \hfil
    \subfloat[DCA \eqref{eq:lse_q1} with $q=1$ versus \eqref{eq:lasso_pf} with $q=2$ (ridge)]{
        \label{subfig:discrete_ridge_vs_sub}
        \includegraphics[width=2.9in]{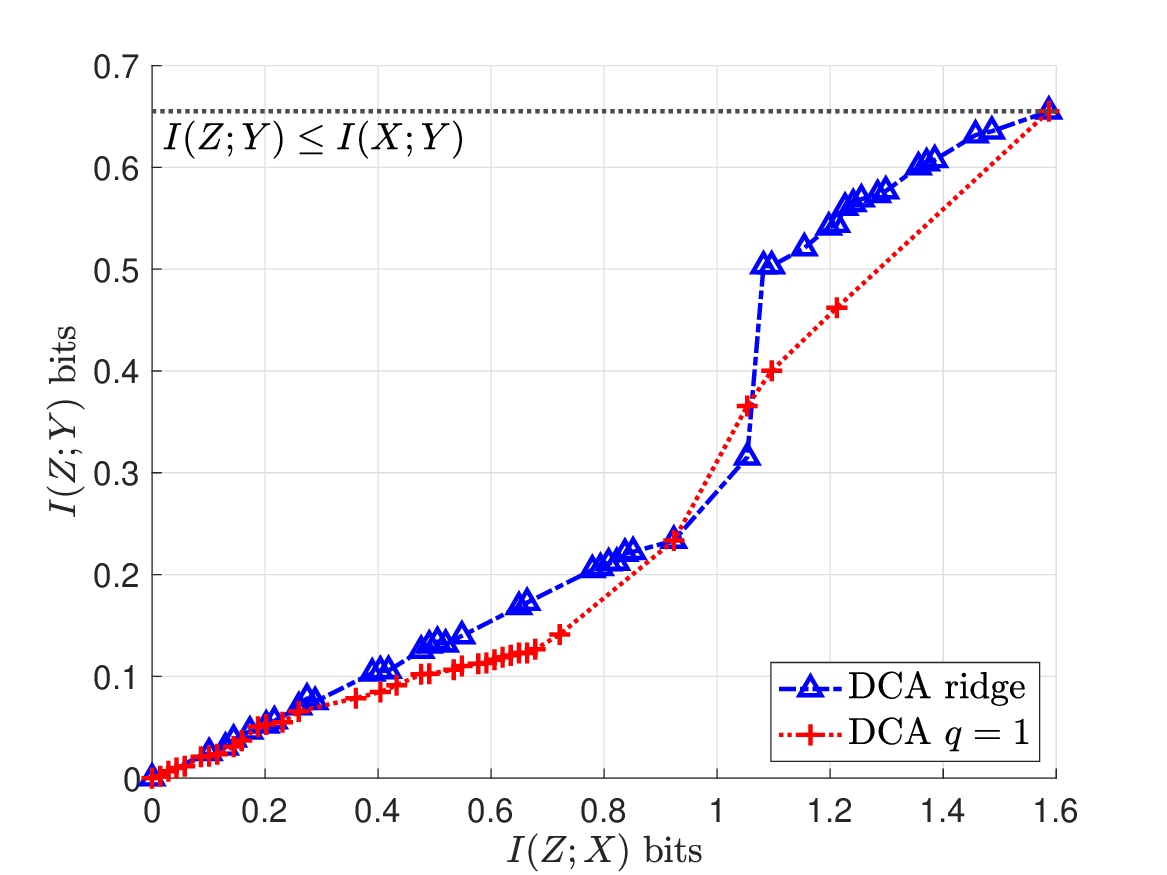}
    }
    \label{fig:discrete_infocurve}
    \caption{Comparing the characterization of the privacy-utility trade-off between the proposed DCA solvers and the state-of-the-art method~\cite{submodpf2019}. The joint distribution $P(X,Y)$ is known and is given by~\eqref{eq:syn_discrete}.}
\end{figure*}
\subsection{Known Distribution}
The joint distribution we consider in this part is given by:
\begin{IEEEeqnarray}{rCl}
    \boldsymbol{P}_{Y|X} = \begin{bmatrix}
        0.90 & 0.08 & 0.40\\
        0.025 & 0.82 & 0.05\\
        0.075 & 0.10 & 0.55
    \end{bmatrix},\quad \boldsymbol{p}_X=\begin{bmatrix}
        \frac{1}{3}\\ \frac{1}{3}\\\frac{1}{3}
    \end{bmatrix}.\IEEEeqnarraynumspace\IEEEyesnumber\label{eq:syn_discrete}
\end{IEEEeqnarray}
Note that in~\eqref{eq:syn_discrete}, $y_0,y_1$ are easy to infer if observing $x_0,x_1$ whereas there is ambiguity in inferring $y_0,y_2$ if $x_2$ is observed instead.

The goal of our evaluation is to characterize the privacy-utility trade-off by plotting the lowest achieved $I(Z;Y)$ for an obtained $I(Z;X)$. The characterized trade-off is displayed on the information plane~\cite{privacyF2014,huang2022linear}, which depends on both $P(X,Y)$ and the algorithm used. We evaluate our DCA solver~\eqref{eq:pf_dca_fg} and compare it to the state-of-the-art solvers~\cite{privacyF2014,submodpf2019}. These baselines only consider deterministic $P(z|x)=\boldsymbol{1}\{x\in z\}$, that is, $z$ categorizes $x,\forall x\in X$. These greedy algorithms merge two or more $z\in\mathcal{Z}$ iteratively for lower PF Lagrangian loss. Contrary to them, our solver is non-greedy with feasible solutions cover the (smoothed) probability simplice. For these baselines, it suffices to compare to~\cite{submodpf2019} since~\cite{privacyF2014} is a special case. 

In the proposed solver~\eqref{eq:pf_dca_fg}, the range of cardinality $|\mathcal{Z}|$ follows the Carath{\'e}odory Theorem, i.e.,  $|\mathcal{Z}|\leq \max\{|\mathcal{X}|,|\mathcal{Y}|\}+1$~\cite{asoodeh2020bottleneck}. For each cardinality, we initialize $P(Z|X)$ by randomly sample a uniformly distributed source in $[0,1]$, followed by normalization to obtain a starting point. Then, we fix a trade-off parameter $\beta\in[0.1,10.0]$ and a regularization coefficient $\alpha\in[0.1,10.0]$, where each set consists of $16$ geometrically spaced points. For each pair of hyperparameters, we run our solver until either the consecutive loss values satisfy $|\mathcal{L}^k-\mathcal{L}^{k+1}|\leq 10^{-6}$ or a maximum number of iterations $K=10^{4}$ is reached. This procedure is repeated for $10$ times. We compute the obtained metrics $I(Z;X),I(Z;Y)$ offline. 

In Fig.~\ref{subfig:discrete_q1_vs_sub}, our solver \eqref{eq:pf_dca_fg} characterizes the trade-off better than the baseline solvers since we not only cover their solutions but achieve more non-trivial points that are infeasible to them. In Fig. \ref{subfig:discrete_ridge_vs_sub} we compare the two proposed DCA solvers with different implementations, i.e.,~\eqref{eq:lasso_pf} with $q=2$ (ridge) versus~\eqref{eq:lse_q1} with $q=1$. While the ridge approach performs worse than the other solver in characterizing the trade-off, it obtains a better trade-off point at $I(Z;X)\approx 1$ bit and $I(Z;Y)\approx 0.3$ bits. Moreover, the ridge solver benefits from the optimized off-the-shelf solver~\cite{scikit-learn}, hence converges faster in our empirical evaluation. It is worth noting that both approaches outperform~\cite{submodpf2019} in characterizing the trade-off under different regularization criterion, demonstrating the strength of our DCA approach.

\subsection{Unknown Distribution}\label{subsec:unkn_eval}
\begin{figure*}[!ht]
    \centering
    \subfloat[MNIST]{
        \includegraphics[width=2.9in]{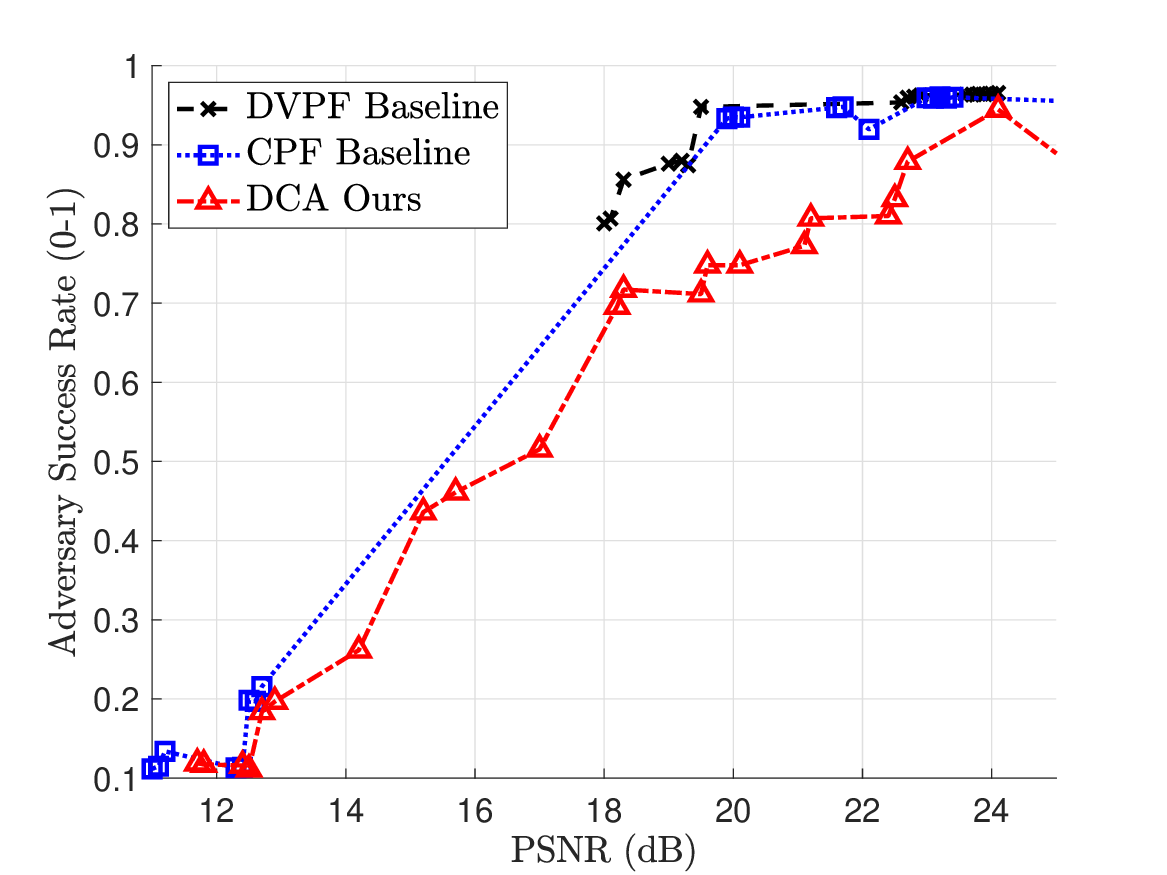}
        \label{subfig:unkn_mnist_all}
    }
    \hfil
    \subfloat[Fashion-MNIST]{
        \includegraphics[width=2.9in]{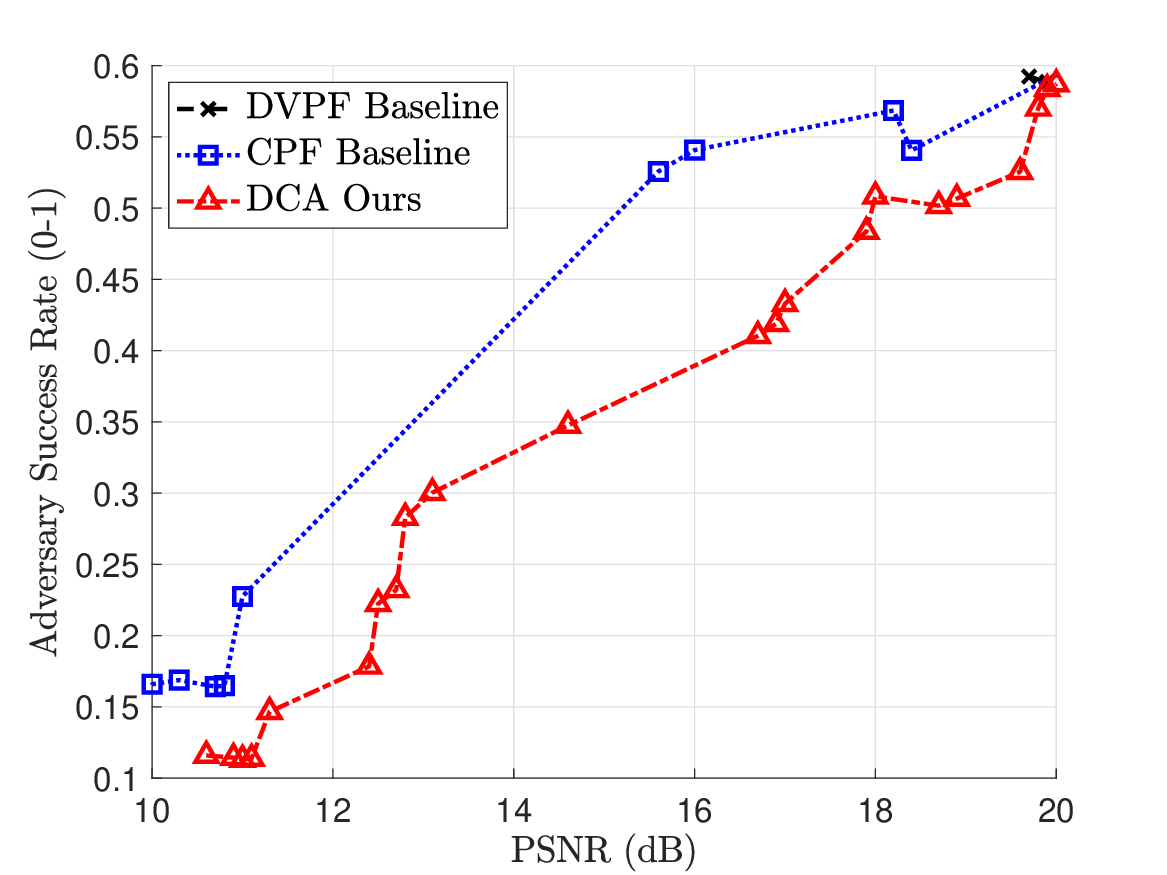}
        \label{subfig:unkn_fashion_all}
    }
    \caption{Privacy-utility trade-off in unknown joint distribution settings. The proposed DCA solver is compared to the CPF~\cite{rodriguez2021variational} and DVPF~\cite{razeghi2024deep} baselines.}
    \label{fig:unknown_eval_all}
\end{figure*}
For unknown $P(X,Y)$, we consider cases where empirical samples of the joint distribution are available (datasets) $\mathcal{D}_N:=\{(x^{(n)},y^{(n)})|(x^{(n)},y^{(n)})\sim P(X,Y)\}_{n=1}^N$. Our DCA solver is compared to the state-of-the-art Conditional Privacy Funnel (CPF)~\cite{rodriguez2021variational} and Deep Variational Privacy Funnel (DVPF)~\cite{razeghi2024deep} baselines.


\paragraph{Datasets}
we evaluate our solver~\eqref{eq:unkn_alg} on MNIST~\cite{lecun2010mnist} and Fashion MNIST~\cite{xiao2017fashion} datasets. Both datasets consist of gray-scale images of $28\times 28$ pixels. Each pixel is normalized to the range $[0,1]$. There are labeled $60000$ training and $10000$ testing samples uniformly distributed between $10$ classes for each dataset. Specifically, the MNIST dataset consists of images of hand-written digits. Here, we let the digits $Y$ be the private information and the images are treated as the public information $X$. As for Fashion MNIST, it has $10$ categories of clothing. We refer to~\cite{xiao2017fashion} for details. We set the categories as $Y$ and the gray-scale images as $X$.

We consider the following scenario: a sender learns (supervised) an encoder-decoder pair with training dataset based on the PF objective. The decoder is shared to a receiver through a secure channel before inference phase. Then the sender encodes $X$ of the testing set as the released codes $Z$. A receiver observes the coded $Z$ and reconstruct $\hat{X}$ with the decoder $P_\phi(X|Z)$. Meanwhile, an adversary also observes the codes but does not know $P_\phi(X|Z)$. The adversary therefore learns $\hat{Y}$ from clustering $Z$. Here, we assume the cardinality of $|Y|$ is known for simplicity (relaxed with cluster identification algorithms~\cite{ray1999determination}). The legitimate users' goal is to reconstruct $\hat{X}$ from $Z$ whereas the adversary clusters $Z$ as $\hat{Y}$ that should reveal $Y$.

In our evaluation, we measure the reconstruction quality (utility) by the Peak Signal-to-Noise power Ratio (PSNR), defined as:$PSNR:=10\log{X_{\max}/MSE}$, where $X_{\max}$ is the maximum possible pixel value ($1$ in our case) and $MSE:=\frac{1}{N}\sum_{i=1}^N\lVert\boldsymbol{x}_i-\hat{\boldsymbol{x}}\rVert_2^2$. As for the privacy leakage, the adversary uses UMAP~\cite{mcinnes2018umap} to project $Z$ to a $\mathbb{R}^2$ space, followed by KMeans clustering~\cite{scikit-learn}. Note that this two-step approach achieves the state-of-the-art clustering performance for both MNIST and Fashion-MNIST datasets~\cite{mcinnes2018umap}. Then we compute the adversary's success rate from offline label matching. 

\paragraph{Network Architecture}
The proposed and compared methods are implemented as deep neural networks. Since our focus is on comparing the PF objective, the encoders and decoders are carefully aligned for all compared methods. We implement the encoder as a neural network with fully connected (FC) layers as the standard VAE~\cite{kingma2013auto} encoder. We have a separate set of parameters for learning \textit{a priori} $P_\theta(Z)=\mathcal{N}(\boldsymbol{0},\boldsymbol{\nu}_\theta)$. As for the decoder, we employ another neural network with FC layers as the reconstruction module. For the differences, our method has a separate linear classifier whereas the CPF baseline expands the decoder's input by concatenating code samples $\boldsymbol{z}$ with the one-hot representation of target variables, i.e., $\boldsymbol{e}_y$ ($\boldsymbol{e}_i$ is the $i^{th}$ elementary vector). For the DVPF baseline, extra adversarial generative networks are implemented according to~\cite{razeghi2024deep}.

We train each model for $100$ epochs with a learning rate at $3\times10^{-4}$ using a standard ADAM optimizer~\cite{kingma2014adam}. The mini-batch size is $256$. For the dimension of $Z$, we vary $d_z\in\{32,64,128,256\}$ for all methods. The range of hyperparameters for our method is set to $[5\times 10^{-4},5\times 10^{-2}]$ for both $\alpha$ and $\beta$ in \eqref{eq:unkn_alg}, whereas it is $[10^{-3},10]$ for the CPF and DVPF baselines. Each hyperparameter pair is trained for $3$ times with a different random seed and we report the average in the results. The experiment details are included in Appendix~\ref{appendix:exp_details}. For reproduction, our source codes in available online\footnote{available at: https://github.com/hui811116/dcaPF-torch}.

\paragraph{Results}
In Fig.~\ref{fig:unknown_eval_all}, we report the highest adversary success rate of an achieved PSNR for each method. Note that the results include all combinations of the hyperparameters and the set of $d_z$. Fig.~\ref{subfig:unkn_mnist_all} corresponds to the MNIST dataset whereas the results for Fashion-MNIST dataset is shown in Fig.~\ref{subfig:unkn_fashion_all}. In both figures, our method clearly outperforms the CPF and DVPF baselines since the adversary has lower success rate in clustering $Z$ from our approach at a comparable reconstruction quality of $X$. This demonstrates that our approach characterizes the privacy-utility trade-off better than both baselines. In addition to better robustness, our approach is more efficient in training phase than the DVPF baseline since we require no parameters for generative and discriminative modules whose optimization follows by a complex six-step algorithm. As for in inference phase, our approach is completely independent to the private information, in sharp contrast to the CPF baseline. For extended evaluation and visualization of the compressed features, we refer to Appendix~\ref{appendix:ex_compress} and Appendix~\ref{appendix:viszz} for details.

\section{Conclusions}
We leverage the DC structure of the PF method and develop efficient solvers based on DCA. In known distribution settings, our solver resembles both ridge regression and sparse recovery. Empirically, our solver outperforms the state-of-the-art greedy solvers in characterizing the privacy-utility trade-off. As for the unknown distribution settings, the insights from the proposed DC approach complies the fundamental Markov relation, so our approach applies to scenarios without labeled private information after training in contrast to compared methods. The evaluation of the proposed solver in MNIST and the Fashion-MNIST datasets demonstrates significantly improved characterization of the privacy-utility trade-off over the compared methods. For future work, we empirically find that the compression ratio ($|\mathcal{Z}|/|\mathcal{X}|$), significantly affects the characterization of the privacy-utility trade-off. A fundamental study in this direction will be the focus of a follow-up work.

\bibliographystyle{ieeetr}
\bibliography{references}

\begin{thebibliography}{10}

\bibitem{privacyF2014}
A.~Makhdoumi, S.~Salamatian, N.~Fawaz, and M.~Médard, ``From the information
  bottleneck to the privacy funnel,'' in {\em 2014 IEEE Information Theory
  Workshop (ITW 2014)}, pp.~501--505, 2014.

\bibitem{cover1999elements}
T.~M. Cover, {\em Elements of information theory}.
\newblock John Wiley \& Sons, 1999.

\bibitem{submodpf2019}
N.~Ding and P.~Sadeghi, ``A submodularity-based clustering algorithm for the
  information bottleneck and privacy funnel,'' in {\em 2019 IEEE Information
  Theory Workshop (ITW)}, pp.~1--5, 2019.

\bibitem{huang2022linear}
T.-H. Huang, A.~E. Gamal, and H.~E. Gamal, ``A linearly convergent
  {D}ouglas-{R}achford splitting solver for {M}arkovian information-theoretic
  optimization problems,'' {\em IEEE Transactions on Information Theory},
  vol.~69, no.~5, pp.~3372--3399, 2023.

\bibitem{attouch2010proximal}
H.~Attouch, J.~Bolte, P.~Redont, and A.~Soubeyran, ``Proximal alternating
  minimization and projection methods for nonconvex problems: An approach based
  on the {K}urdyka-{\l}ojasiewicz inequality,'' {\em Mathematics of operations
  research}, vol.~35, no.~2, pp.~438--457, 2010.

\bibitem{alemi2016deep}
A.~A. Alemi, I.~Fischer, J.~V. Dillon, and K.~Murphy, ``Deep variational
  information bottleneck,'' {\em arXiv preprint arXiv:1612.00410}, 2016.

\bibitem{rodriguez2021variational}
B.~Rodr{\'\i}guez-G{\'a}lvez, R.~Thobaben, and M.~Skoglund, ``A variational
  approach to privacy and fairness,'' in {\em 2021 IEEE Information Theory
  Workshop (ITW)}, pp.~1--6, IEEE, 2021.

\bibitem{razeghi2024deep}
B.~Razeghi, P.~Rahimi, and S.~Marcel, ``Deep variational privacy funnel:
  General modeling with applications in face recognition,'' {\em arXiv preprint
  arXiv:2401.14792}, 2024.

\bibitem{le2018dc}
H.~A. Le~Thi and T.~Pham~Dinh, ``Dc programming and dca: thirty years of
  developments,'' {\em Mathematical Programming}, vol.~169, no.~1, pp.~5--68,
  2018.

\bibitem{7282765}
F.~P. Calmon, A.~Makhdoumi, and M.~Médard, ``Fundamental limits of perfect
  privacy,'' in {\em 2015 IEEE International Symposium on Information Theory
  (ISIT)}, pp.~1796--1800, 2015.

\bibitem{6942222}
C.~Schieler and P.~Cuff, ``Rate-distortion theory for secrecy systems,'' {\em
  IEEE Transactions on Information Theory}, vol.~60, no.~12, pp.~7584--7605,
  2014.

\bibitem{7541465}
K.~Kittichokechai and G.~Caire, ``Privacy-constrained remote source coding,''
  in {\em 2016 IEEE International Symposium on Information Theory (ISIT)},
  pp.~1078--1082, 2016.

\bibitem{9678374}
Y.~Yakimenka, H.-Y. Lin, E.~Rosnes, and J.~Kliewer, ``Optimal
  rate-distortion-leakage tradeoff for single-server information retrieval,''
  {\em IEEE Journal on Selected Areas in Communications}, vol.~40, no.~3,
  pp.~832--846, 2022.

\bibitem{9272984}
Y.~Y. Shkel, R.~S. Blum, and H.~V. Poor, ``Secrecy by design with applications
  to privacy and compression,'' {\em IEEE Transactions on Information Theory},
  vol.~67, no.~2, pp.~824--843, 2021.

\bibitem{donoho2006compressed}
D.~L. Donoho, ``Compressed sensing,'' {\em IEEE Transactions on information
  theory}, vol.~52, no.~4, pp.~1289--1306, 2006.

\bibitem{draper2002generalized}
N.~R. Draper and F.~Pukelsheim, ``Generalized ridge analysis under linear
  restrictions, with particular applications to mixture experiments problems,''
  {\em Technometrics}, vol.~44, no.~3, pp.~250--259, 2002.

\bibitem{gaines2018algorithms}
B.~R. Gaines, J.~Kim, and H.~Zhou, ``Algorithms for fitting the constrained
  lasso,'' {\em Journal of Computational and Graphical Statistics}, vol.~27,
  no.~4, pp.~861--871, 2018.

\bibitem{pilanci2012recovery}
M.~Pilanci, L.~Ghaoui, and V.~Chandrasekaran, ``Recovery of sparse probability
  measures via convex programming,'' {\em Advances in Neural Information
  Processing Systems}, vol.~25, 2012.

\bibitem{scikit-learn}
F.~Pedregosa, G.~Varoquaux, A.~Gramfort, V.~Michel, B.~Thirion, O.~Grisel,
  M.~Blondel, P.~Prettenhofer, R.~Weiss, V.~Dubourg, J.~Vanderplas, A.~Passos,
  D.~Cournapeau, M.~Brucher, M.~Perrot, and E.~Duchesnay, ``Scikit-learn:
  Machine learning in {P}ython,'' {\em Journal of Machine Learning Research},
  vol.~12, pp.~2825--2830, 2011.

\bibitem{kingma2013auto}
D.~P. Kingma and M.~Welling, ``Auto-encoding variational {B}ayes,'' {\em arXiv
  preprint arXiv:1312.6114}, 2013.

\bibitem{duchi2007derivations}
J.~Duchi, ``Derivations for linear algebra and optimization,'' {\em Berkeley,
  California}, vol.~3, no.~1, pp.~2325--5870, 2007.

\bibitem{park2019variational}
Y.~Park, C.~Kim, and G.~Kim, ``Variational {L}aplace autoencoders,'' in {\em
  International conference on machine learning}, pp.~5032--5041, PMLR, 2019.

\bibitem{huang2023Wynerfinger}
T.-H. Huang, T.~Dahanayaka, K.~Thilakarathna, P.~H. Leong, and H.~El~Gamal,
  ``The {W}yner variational autoencoder for unsupervised multi-layer wireless
  fingerprinting,'' in {\em GLOBECOM 2023 - 2023 IEEE Global Communications
  Conference}, pp.~820--825, 2023.

\bibitem{asoodeh2020bottleneck}
S.~Asoodeh and F.~P. Calmon, ``Bottleneck problems: An information and
  estimation-theoretic view,'' {\em Entropy}, vol.~22, no.~11, p.~1325, 2020.

\bibitem{lecun2010mnist}
Y.~LeCun, C.~Cortes, and C.~Burges, ``Mnist handwritten digit database,'' {\em
  ATT Labs [Online]. Available: http://yann.lecun.com/exdb/mnist}, vol.~2,
  2010.

\bibitem{xiao2017fashion}
H.~Xiao, K.~Rasul, and R.~Vollgraf, ``Fashion-mnist: a novel image dataset for
  benchmarking machine learning algorithms,'' {\em arXiv preprint
  arXiv:1708.07747}, 2017.

\bibitem{ray1999determination}
S.~Ray and R.~H. Turi, ``Determination of number of clusters in k-means
  clustering and application in colour image segmentation,'' in {\em
  Proceedings of the 4th international conference on advances in pattern
  recognition and digital techniques}, vol.~137, p.~143, Calcutta, India:,
  1999.

\bibitem{mcinnes2018umap}
L.~McInnes, J.~Healy, and J.~Melville, ``Umap: Uniform manifold approximation
  and projection for dimension reduction,'' {\em arXiv preprint
  arXiv:1802.03426}, 2018.

\bibitem{kingma2014adam}
D.~P. Kingma and J.~Ba, ``{ADAM}: A method for stochastic optimization,'' {\em
  arXiv preprint arXiv:1412.6980}, 2014.

\bibitem{10206810}
T.-H. Huang and H.~El~Gamal, ``Efficient alternating minimization solvers for
  {W}yner multi-view unsupervised learning,'' in {\em 2023 IEEE International
  Symposium on Information Theory (ISIT)}, pp.~707--712, 2023.

\bibitem{huang2024efficient}
T.-H. Huang and H.~E. Gamal, ``Efficient solvers for {W}yner common information
  with application to multi-modal clustering,'' {\em arXiv preprint
  arXiv:2402.14266}, 2024.

\bibitem{radford2015unsupervised}
A.~Radford, L.~Metz, and S.~Chintala, ``Unsupervised representation learning
  with deep convolutional generative adversarial networks,'' {\em arXiv
  preprint arXiv:1511.06434}, 2015.

\end{thebibliography}

\appendices

\section{Derivation for a known joint distribution}\label{appendix:derive_known_jt}
Starting from \eqref{eq:pf_dca_fg}:
\begin{IEEEeqnarray}{rCl}
    f(\boldsymbol{p}_{z|x}):&=&-H(Z|Y),\nonumber\\
    g(\boldsymbol{p}_{z|x}):&=&-H(Z)+\beta I(Z|X),\nonumber
\end{IEEEeqnarray}
we note that the ``linearized'' part has the following closed-form expression from elementary functional derivative:
\begin{IEEEeqnarray}{rCl}
    \frac{\partial g(\boldsymbol{p}_{z|x})}{\partial P(z|x)}:=P(x)\left\{\log{P(z)}+1+\beta \log{\frac{P(z|x)}{P(z)}}\right\}.\nonumber
\end{IEEEeqnarray}
Furthermore, when analyzing the first-order necessary condition of the right-hand-side of \eqref{eq:DCA_for_pf}, we have:
\begin{IEEEeqnarray}{rCl}
    \sum_{y\in\mathcal{Y}}P(y|x)\log{P(z|y)}=\left[\log{P^k(z)}+\beta\log{\frac{P^k(z|x)}{P^k(z)}}\right],\nonumber
\end{IEEEeqnarray}
By definition, $\boldsymbol{B}_{y|x}:=\boldsymbol{I}_{|\mathcal{Z}|}\otimes \boldsymbol{Q}_{y|x}$ where $\boldsymbol{Q}_{y|x}$ is the matrix form of $P(Y|X)$ with $(i,j)$-entry $P(y_i|x_j)$, so the above is equivalent to:
\begin{IEEEeqnarray}{rCl}
    \boldsymbol{B}_{y|x}\log{\boldsymbol{p}_{z|y}}=\log{\boldsymbol{p}^k_z}+\beta\log{\frac{\boldsymbol{p}^k_{z|x}}{\boldsymbol{p}^k_z}}.\nonumber
\end{IEEEeqnarray}

\section{Alternative Approach for Sparse Recovery}\label{appendix:kn_sparse_problem}
The transformation of the optimization problem~\eqref{eq:inner_update_opt_problem} for the case $q=1$ to~\eqref{eq:lse_q1} is achieved through the following steps:

First, we change the variables from the conditional probabilities $\boldsymbol{p}_{z|x}$ to the log-likelihoods $l_{z|x}:=\log{\boldsymbol{p}_{z|x}}$. The benefit of this is that the domain becomes a half-space $l_{z|x}\in(-\infty,0]$, which is more straightforward for projection than the probability simplice for $\boldsymbol{p}_{z|x}$ as identified in~\cite{10206810}. Second, instead of finding the exact deterministic assignment, i.e., $P(z|x)=\boldsymbol{1}\{x\in\mathcal{Z}\}$, we impose smoothness conditions on the log-likelihoods, $-M\leq l_{z|x}\leq -m$ for some $M,m>0$. Lastly, we compute the Markov relation for $l_{z|x}$ as:
\begin{IEEEeqnarray}{rCl}
    \log{\boldsymbol{p}_{z|y}}=\log\sum_{x\in\mathcal{X}}\exp\{l_{x|y}+l_{z|x}\},\nonumber
\end{IEEEeqnarray}
where $l_{x|y}:=\log{\boldsymbol{p}_{x|y}}$. The above is implemented with a log-sum-exponential function $lse(x):=\log\sum_{x\in\mathcal{X}}\exp\{x\}$.
Combining the above, we rewrite \eqref{eq:lasso_pf} ($q=1$) as:
\begin{IEEEeqnarray}{rCl}
    l^{k+1}_{z|x}:=\underset{l_{z|x}\in \mathcal{H}_{z|x}}{\arg\min}\, \frac{1}{2}\lVert {lse}_{x}(l_{x|y}+l_{z|x})-l^k_{\phi}\rVert^2+\alpha\lVert l_{z|x}\rVert_1.\nonumber
\end{IEEEeqnarray}
We therefore complete the transformation.

\section{Derivation for unknown joint distribution}\label{appendix:derive_unkn_jt}
Again, we start with the update equation:
\begin{IEEEeqnarray}{rCl}
    \sum_{y\in\mathcal{Y}}P(y|x)\log{P(z|y)}=\left[\log{P^k(z)}+\beta\log{\frac{P^k(z|x)}{P^k(z)}}\right],\nonumber
\end{IEEEeqnarray}
but now we take expectation with respect to $P(Z,X)=P(X)P(Z|X)$. The left-hand side is:
\begin{IEEEeqnarray}{rCl}
    &&\mathbb{E}_{z,x}\left[\sum_{y\in\mathcal{Y}}P(y|X)\log{P(Z|y)}\right]\nonumber\\
    =&&\mathbb{E}_{y}\left[\sum_{z\in\mathcal{Z}}\sum_{x\in\mathcal{X}}P(z,x)P(x|Y)\log{P(z|Y)}\right]\nonumber\\
    =&&-H(Z|Y),\nonumber
\end{IEEEeqnarray}
where the second equality is due to the Markov relation $Y\rightarrow X\rightarrow Z$. For the second term of~\eqref{eq:expect_dca_update}, we have:
\begin{IEEEeqnarray}{rCl}
    \mathbb{E}_{z}\left[\log{P^k(Z)}\right]=-H(Z)-D_{KL}\left[P_z\parallel P^k_z\right].
\end{IEEEeqnarray}
Combining the above with some re-arrangement, we have:
\begin{IEEEeqnarray}{rCl}
    I(Z;Y)+D_{KL}[P_z\parallel P_z^k]-\beta\mathbb{E}_{z,x}\left[\log{P^k(X|Z)}\right]=0.\nonumber
\end{IEEEeqnarray}
This completes the proof.

\section{Proof of Theorem~\ref{thm:conv_discrete_kn}}\label{appendix:proof_thm}
The proof follows that in~\cite[Theorem 1]{huang2024efficient}. Which is included here for completeness.

First we establish the following lemma, using the fact that the linear equation $\boldsymbol{p}_{z}=\boldsymbol{A}_x\boldsymbol{p}_{z|x}$ represents $P(Z)=\sum_{x\in\mathcal{X}}P(X)P(Z|X)$, where $\boldsymbol{A}_x:=\boldsymbol{I}_{|\mathcal{Z}|}\otimes \boldsymbol{p}_x$ denotes the matrix form of the known marginal $P(X)$:
\begin{lemma}[Restricted Convexity]\label{lemma:res_cvx_g}
    Let $g(\boldsymbol{p}_{z|x})$ be defined as in~\eqref{eq:pf_dca_fg}, then we have:
    \begin{IEEEeqnarray*}{rCl}
        g(\boldsymbol{p}_{z|x})&\geq& g(\boldsymbol{q}_{z|x})+\langle\nabla g(\boldsymbol{q}_{z|x}),\boldsymbol{p}_{z|x}-\boldsymbol{q}_{z|x}\rangle\nonumber\\
        &&+\frac{1}{2}\lVert \boldsymbol{A}_x(\boldsymbol{p}_{z|x}-\boldsymbol{q}_{z|x})\rVert^2_2,
    \end{IEEEeqnarray*}
    where $\boldsymbol{p}_{z|x},\boldsymbol{q}_{z|x}\in\Omega_{z|x}$, with $\Omega_{z|x}$ denotes the feasible solution set of the conditional probability $P(Z|X)$ in vector form.
\end{lemma}
\begin{IEEEproof}
    It is well-known that the negative Shannon entropy $-H(X):=-H(\boldsymbol{p}_x)$ is strongly convex with respect to the probability vector~$\boldsymbol{p}_x$. Observe that $-H(Z)$ is a component of $g(\boldsymbol{p}_{z|x})$ and that the marginal $\boldsymbol{p}_z=\boldsymbol{A}_x\boldsymbol{p}_{z|x}$, and that the other components are also convex. The proof is complete.
\end{IEEEproof}

By the first order necessary condition of the proposed DCA solver~\eqref{eq:DCA_for_pf}, we have:
\begin{IEEEeqnarray}{rCl}
    \nabla f(\boldsymbol{p}_{z|x}^{k+1}) -\nabla g(\boldsymbol{p}_{z|x}^k) =0.\IEEEeqnarraynumspace\IEEEyesnumber\label{eq:appendix_fg_fonc}
\end{IEEEeqnarray}
Denote the objective function as $\mathcal{L}(\boldsymbol{p})=f(\boldsymbol{p})-g(\boldsymbol{p})$ where $f,g$ are convex functions of $\boldsymbol{p}\in\Omega_{z|x}$. We consider the following:
\begin{IEEEeqnarray}{rCl}
    &&\mathcal{L}(\boldsymbol{p}_{z|x}^k)-\mathcal{L}(\boldsymbol{p}_{z|x}^{k+1})\nonumber\\
    &=& f(\boldsymbol{p}_{z|x}^k)-g(\boldsymbol{p}_{z|x}^k)-f(\boldsymbol{p}_{z|x}^{k+1})+g(\boldsymbol{p}_{z|x}^{k+1})\nonumber\\
    &\geq& \langle\nabla f(\boldsymbol{p}_{z|x}^{k+1})-\nabla g(\boldsymbol{p}_{z|x}^k),\boldsymbol{p}_{z|x}^k-\boldsymbol{p}_{z|x}^{k+1}\rangle\nonumber\\
    &&+\frac{1}{2}\lVert \boldsymbol{A}_{x}(\boldsymbol{p}_{z|x}^k-\boldsymbol{p}_{z|x}^{k+1})\rVert^2_2,\nonumber
\end{IEEEeqnarray}
where the first inequality follows from the convexity of $f$, followed by the restricted convexity of $g$ as shown in Lemma~\ref{lemma:res_cvx_g}. Then by applying~\eqref{eq:appendix_fg_fonc}, we show that:
\begin{IEEEeqnarray}{rCl}
    \mathcal{L}(p^k)-\mathcal{L}(p^{k+1})\geq \frac{1}{2}\lVert\boldsymbol{A}_x(\boldsymbol{p}^k_{z|x}-\boldsymbol{p}_{z|x}^{k+1})\rVert_2^2.
\end{IEEEeqnarray}
The above implies that the consecutive updates from~\eqref{eq:DCA_for_pf} results in a monotone decreasing sequence of the objective values. Summing both side from $k=0$ to $k=K-1$ sufficiently large, we have:
\begin{IEEEeqnarray*}{rCl}
    \mathcal{L}(\boldsymbol{p}^0_{z|x})-\mathcal{L}(\boldsymbol{p}^K_{z|x})\geq \sum_{k=0}^{K-1}\lVert \boldsymbol{A}_x(\boldsymbol{p}^k_{z|x}-\boldsymbol{p}_{z|x}^{k+1})\rVert^2_2.
\end{IEEEeqnarray*}
For $K\rightarrow \infty$ sufficiently large, and under mild smoothness conditions~\cite{huang2022linear}, $\mathcal{L}$ is lower-semi-continuous, and hence the left hand side is finite, which in turns implies that the Cauchy sequence in the right hand side is bounded. Therefore, we have a sequence $\{\boldsymbol{A}_x\boldsymbol{p}_{z|x}^k\}_{k\geq K_0}$, obtained from the DCA algorithm~\eqref{eq:DCA_for_pf} such that $\{\mathcal{L}(\boldsymbol{p}_{z|x}^k)\}_{k\geq K_0}$ converges to a $\mathcal{L}^*$ as $k\rightarrow \infty$. Denote the converged solutions $\boldsymbol{A}_x\boldsymbol{p}_{z|x}^*\in\Omega^*$ where $\Omega^*:=\{\boldsymbol{A}_x\boldsymbol{q}|\nabla f(\boldsymbol{q})=\nabla g(\boldsymbol{q}),\boldsymbol{q}\in\Omega_{z|x}\}$, then $\Omega^*$ is a set of stationary points whose objective value is $\mathcal{L}^*$. Now, because $\mathcal{L}$ is non-convex, $\Omega^*$ could include both local and global stationary points.

The above results do not directly apply to the proposed DCA solver because here we have two intermediate optimization problems~\eqref{eq:lasso_pf} with $q=2$ and~\eqref{eq:lse_q1}. Interestingly, both~\eqref{eq:lasso_pf} and \eqref{eq:lse_q1} are convex optimization problems, since the log-sum-exponential function and the $q$-norms for $q\in\{1,2\}$ are all convex functions. Therefore, the convexity assures convergence to the global optimum for the intermediate problems. Combining the above, the proof is complete.

\section{Experiment Details}\label{appendix:exp_details}
We implement all compared solvers in Section~\ref{subsec:unkn_eval} as follows. For the hardware, we run our prototype on a commercial computer with an Intel i5-9400 CPU and one Nvidia GeForce GTX 1650 Super GPU with 16GB memory. As for the software, we implement all solvers with PyTorch. All methods are optimized with a standard ADAM optimizer~\cite{kingma2014adam} with a fixed learning rate of $3\times 10^{-4}$ for $100$ epochs, and we configure the mini-batch size to $256$. In Table~\ref{table:para_vs_codelen} we list the number of parameters of the proposed solver with different dimensionality of $Z$.

\begin{table}[!ht]
\caption{Number of Trainable Parameters versus the code length.}
\label{table:para_vs_codelen}
\centering
\begin{tabular}{cc}
\hline
\textbf{$d_z$} & \textbf{Number of Parameters} \\ \hline\hline
$32$ & $1084742$ \\ \hline
$64$ & $1133190$ \\ \hline
$128$ & $1230086$ \\ \hline
$256$ & $1423878$ \\ \hline
\end{tabular}
\end{table}
As for the off-the-shelf software: UMAP and KMeans, the default configuration is adopted. We refer to the references for details~\cite{mcinnes2018umap,scikit-learn}.

 For both the proposed and compared methods, the deep neural network architectures are described as follows: the encoder is implemented as a multi-layer perceptron (MLP) with $784\rightarrow 500\rightarrow 500\rightarrow (d,d)$ fully-connected (FC) layers where the last pair corresponds to the parameterized mean vector and diagonal covariance matrix of the coded Gaussian vector $\boldsymbol{z}$ similar to the VAE~\cite{kingma2013auto}. As for the decoder, our method consists of two parts: 1) the reconstruction block which reverses the architecture of the encoder but starts with $d$ neurons instead. 2) classifier of the coded $\boldsymbol{z}\sim \boldsymbol{\mu}_x+\boldsymbol{\varepsilon}\circ \boldsymbol{\sigma}_x$ (element-wise), where $\boldsymbol{\varepsilon}\sim\mathcal{N}(\boldsymbol{0},\boldsymbol{I}_d)$. The linear classifier is implemented as a MLP with $d\rightarrow |Y|$ FC layers, followed by a softmax output activation. 
 
 For the compared methods, we follow the implementation details in~\cite{rodriguez2021variational,razeghi2024deep} with minimal modifications. For the CPF baseline, the same encoder is adopted, but the decoder is modified to a $(|\mathcal{Y}|+d)\rightarrow 500\rightarrow 784$ FC MLP whose inputs are concatenated from $\boldsymbol{z}$ and a one-hot label vector (an elementary vector of $|Y|$ dimensions). As for the DVPF baseline, we follow the six-step optimization algorithm~\cite{razeghi2024deep} and implements extra discriminative and generative modules. The generator is a MLP with $100\rightarrow 200\rightarrow d_z$ FC layers with $\text{BatchNorm2d}+\text{LeakyReLU}$ activation pair for the hidden layer as in~\cite{radford2015unsupervised}. Then three discriminators for each of $X,Y,Z$ are implemented as three separate MLPs with $d_w\rightarrow 200\rightarrow 200$ where $w\in\{x,y,z\}$ with the same $\text{BatchNorm2d}+\text{LeakyReLU}$ activation pair for the hidden layer, followed by a $\text{Sigmoid}$ function as the output activation. The training of the generative and discriminative modules carefully follows the algorithm provided in~\cite{razeghi2024deep}.
\begin{figure}[!t]
    \centering
    \includegraphics[width=2.9in]{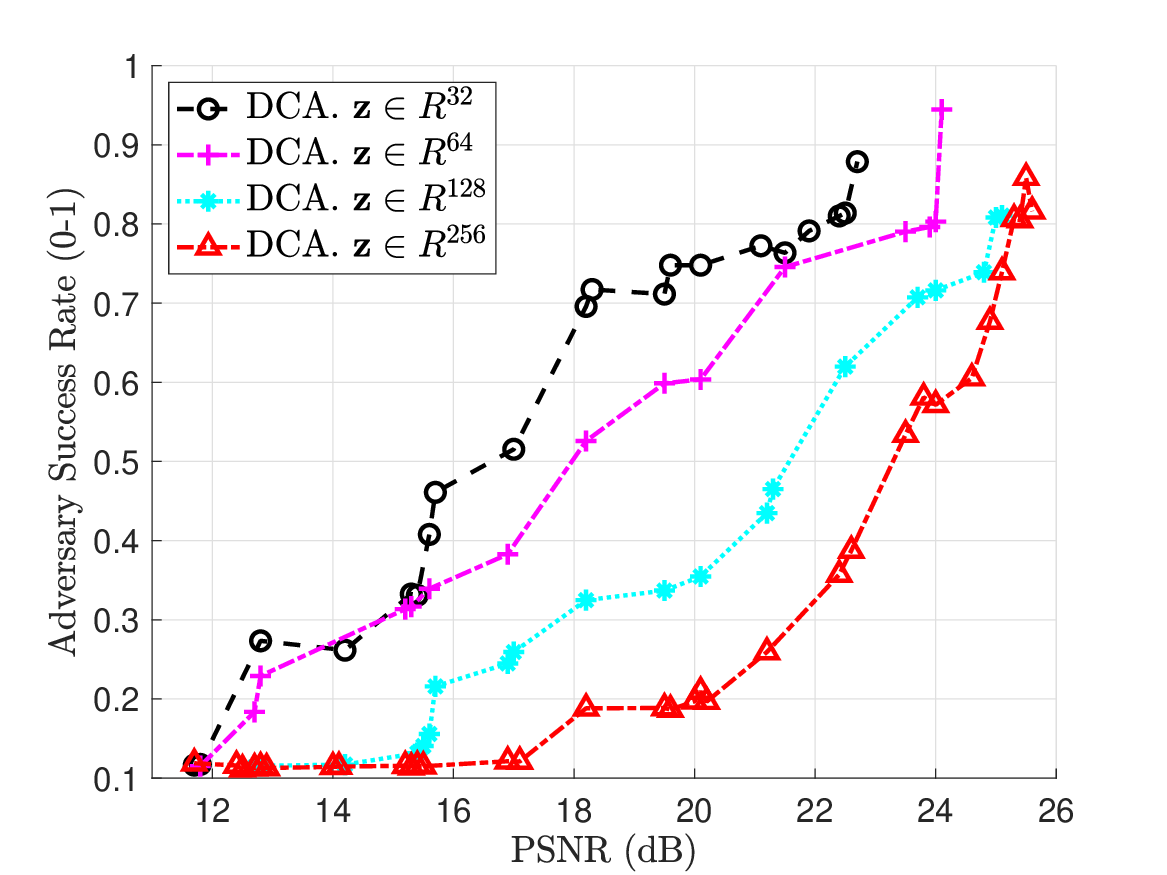}
    \caption{Privacy-utility trade-off of the MNIST dataset in different code length. Here, $\boldsymbol{x}\in\mathbb{R}^{784}$ whereas $\boldsymbol{z}\in\mathbb{R}^d$. }
    \label{fig:appendix_mnist_compress}
\end{figure}
\begin{figure*}[t]
    \centering
    \subfloat[Data Samples]{
        \label{subfig:mnist_data_sample}
        \includegraphics[width=3.2in]{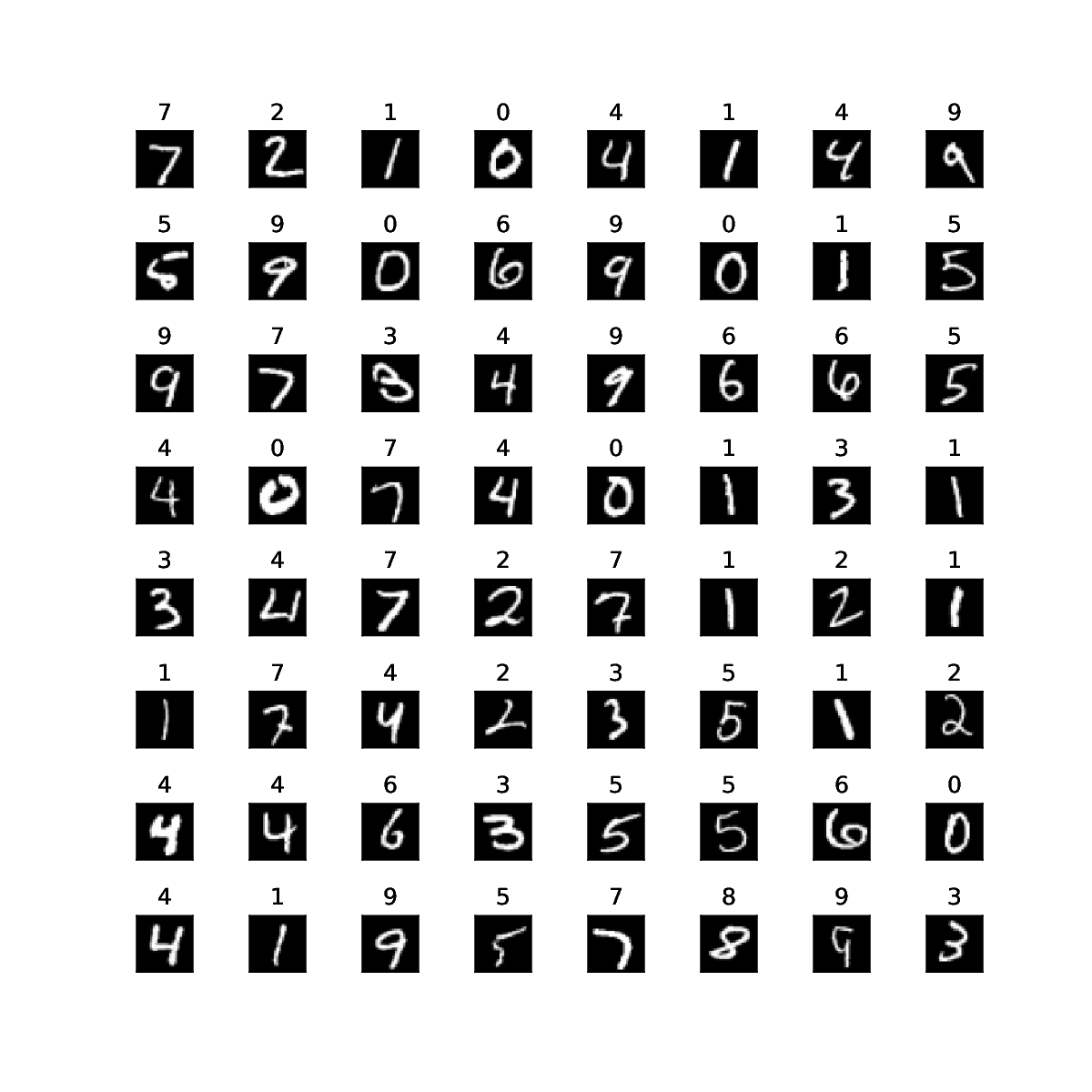}
    }
    \hfil
    \subfloat[Reconstruction (our DCA)]{
        \label{subfig:mnist_reconstruct}
        \includegraphics[width=3.2in]{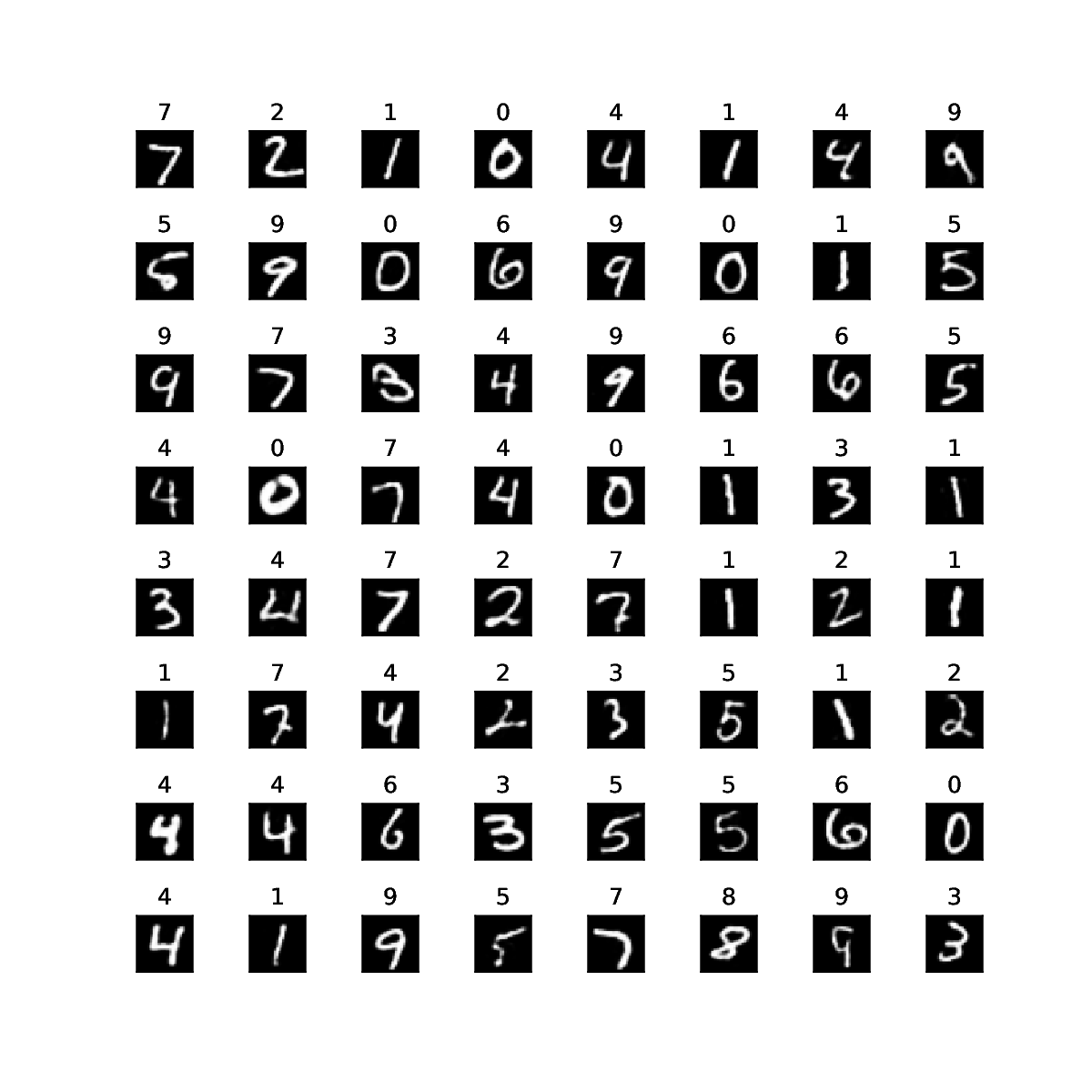}
    }
    \caption{Testing data versus reconstruction from the DCA solver. PSNR=$24$ dB, Adversary success rate=$0.57$}%
    \label{fig:appendix_mnist_samples}%
\end{figure*}
In training the proposed neural networks prototype, we update the weights alternatingly. We fix the weights of the encoder and update the decoder with~\eqref{eq:unkn_alg_step1}. Then we fix the weights of the updated decoder and update the encoder parameters (including that of the prior) through~\eqref{eq:unkn_alg_step2}. This alternating procedure is implemented in PyTorch with the combination of an on/off $\textit{requires\textunderscore grad}$ flag for the parameters of the encoder/decoder and activating the $\textit{retain\textunderscore graph}$ flag in the backpropagation of~\eqref{eq:unkn_alg_step1}. 
\begin{figure*}[t]
    \centering
    \subfloat[Projection with $\boldsymbol{x}$ directly]{
        \label{subfig:vis_z_umap_direct}
        \includegraphics[width=3.2in]{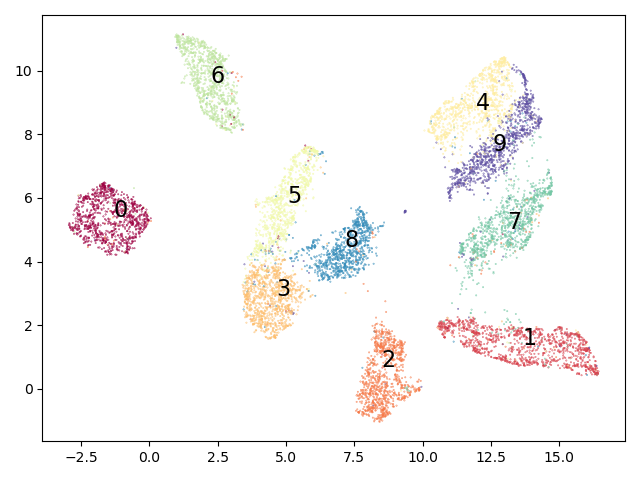}
    }
    \hfil
    \subfloat[Projection with our $\boldsymbol{z}$]{
        \label{subfig:vis_z_dcapf}
        \includegraphics[width=3.2in]{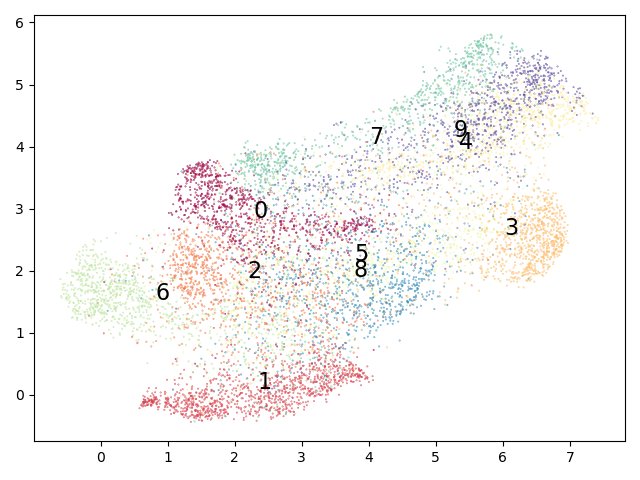}
    }
    \label{fig:appendix_dcapf_code}
    \caption{Visualization of the codes from DCA solver. PSNR=$24$dB, Adversary success rate=$0.57$ and code length $d_z=256$. Centers of each class are labeled.}
\end{figure*}

\section{Extended Evaluation: Compression Ratio and Privacy-Utility Trade-off}\label{appendix:ex_compress}
For the proposed approach, we further evaluate the relationship between the dimensionality of the compressed codes and the privacy-utility trade-off. Here, we focus on the MNIST dataset for its simplicity as shown in Section \ref{subsec:unkn_eval}. Fig. \ref{subfig:unkn_mnist_all} in the main content accounts for all $d_z\in\{32,64,128,256\}$, but here we separate each of the configurations to investigate its effects on the trade-off.

In Fig.~\ref{fig:appendix_mnist_compress}, we show the adversary success rate versus the achieved PSNR of the proposed DCA solver with different configurations of the dimensionalty of $Z$, i.e., $d_z$. Our results show that the DCA solver can be more robust against privacy leakage (higher adversary success rate) with larger $d_z$. For MNIST dataset the number of dimensions of the public information $X$ is $d_x=784$ and since it is often desirable to find a compressed representation of $X$. Therefore, one can readily identify a computational trade-off in parallel to the privacy-utility trade-off. Specifically, a low compression ratio $d_z/d_x$ results in a low robustness against an adversary of the leakage of privacy $Y$ for a fixed reconstruction quality of the public information $X$. The characterization of the computation trade-off will be left for future work.

\section{Extended Evaluation: Visualization}\label{appendix:viszz}
In Fig.~\ref{fig:appendix_mnist_samples}, we show the first $64$ samples of the MNIST testing samples with the reconstruction of the proposed DCA solver with $\alpha=0.01$ and $\beta=0.05$. The PSNR for this configuration results is $24$ dB with adversarial success rate of $0.57$. The samples are perceptually recognizable after reconstruction, demonstrating sufficient utility of the images $X$. Then in Fig.~\ref{subfig:vis_z_umap_direct} and~\ref{subfig:vis_z_dcapf}, we visualize the projected $Z$ of the same configuration. We follow the same method described in Section~\ref{subsec:unkn_eval} and use UMAP~\cite{mcinnes2018umap} to project the representation $Z$ to a $\mathbb{R}^2$ space with the default hyperparameters adopted. Moreover, we compare the $2$-dimensional projection from our code to direct projection of $X$, i.e., no privatization. As shown from Fig.~\ref{subfig:vis_z_dcapf}, projecting our privatized codes has more ambiguity between all classes whereas in Fig.~\ref{subfig:vis_z_umap_direct}, the direct projection baseline reveals recognizable separation between them instead. This means lower privacy leakage for the proposed approach under sufficient reconstruction quality.

\end{document}